\documentclass[journal, doublecolumn]{IEEEtran}
\usepackage{graphicx}
\usepackage{lineno}
\usepackage{subcaption}
\usepackage{amsmath}

\usepackage[table]{xcolor}

\definecolor{myred}{rgb}{1, 0, 0}

\newcommand{\ical}[1]{{\mbox{\usefont{OT1}{pzc}{m}{it}{#1}}}}

\newcommand{\m}[1]{{\mbox{{\fontencoding{T1}\sffamily\slshape{#1\/}}}}}

\begin{document}

\title{Leveraging Activation Maximization and Generative Adversarial Training to Recognize and Explain Patterns in Natural Areas in Satellite Imagery}

\author{Ahmed~Emam~\IEEEmembership{,}
        Timo.~T.~Stomberg~\IEEEmembership{,}
        and~Ribana~Roscher,~\IEEEmembership{Member,~IEEE}
\thanks{A. Emam and T. T. Stomberg are with the Institute
of Geodesy and Geoinformation, University of Bonn, 53113 Bonn, Germany. (aemam,~timo.stomberg)@uni-bonn.de}
\thanks{R. Roscher is with the Institute of Bio- and Geosciences, Forschungszentrum Jülich, 52428 Jülich, Germany, and the Institute
of Geodesy and Geoinformation, University of Bonn, 53113 Bonn, Germany.~(r.roscher@fz-juelich.de)}
\thanks{Manuscript received; revised }}

\markboth{Journal of \LaTeX\ Class Files,~Vol.~, No.~, February~2023}%
{Emam \MakeLowercase{\textit{et al.}}: Leveraging Activation Maximization and Generative Adversarial Training to Recognize and Explain Patterns in Natural Areas in Satellite Imagery}

\maketitle

\begin{abstract}
Natural protected areas are vital for biodiversity, climate change mitigation, and supporting ecological processes. Despite their significance, comprehensive mapping is hindered by a lack of understanding of their characteristics and a missing land cover class definition. This paper aims to advance the explanation of the designating patterns forming protected and wild areas. To this end, we propose a novel framework that uses activation maximization and a generative adversarial model. With this, we aim to generate satellite images that, in combination with domain knowledge, are capable of offering complete and valid explanations for the spatial and spectral patterns that define the natural authenticity of these regions.  Our proposed framework produces more precise attribution maps pinpointing the designating patterns forming the natural authenticity of protected areas.  Our approach fosters our understanding of the ecological integrity of the protected natural areas and may contribute to future monitoring and preservation efforts.
\end{abstract}


\IEEEpeerreviewmaketitle

\section{Introduction}

\IEEEPARstart{N}{atural} areas with minimum human influence refer to regions on Earth that remain largely untouched by human activities such as urbanization, agriculture, logging, and other forms of development.
These areas are characterized by high levels of biodiversity and provide a wealth of ecological integrity benefits; for example, these areas offer unique access to understanding nearly undisrupted natural ecosystem processes such as water and pollination cycles. Therefore, comprehensive mapping and monitoring of highly protected natural areas are crucial to understanding the geo-ecological patterns essential for these areas to thrive \cite{Sarkar1999WildernessPA}.
With this in mind, it is not surprising that the monitoring and understanding of natural areas have gained significant attention in recent years within the remote sensing and ecological research communities.

Satellite imagery presents an effective approach for continuously monitoring vast protected natural areas that are challenging for humans to access. By utilizing this technology, efficient and cost-effective data collection becomes possible while minimizing disturbances to sensitive ecosystems. Leveraging large satellite imagery datasets, machine learning models like deep convolutional neural networks (CNN) can accurately differentiate natural areas. For instance, in a study conducted by the authors of \cite{ekim_mapinwild_2023}, a dataset and a baseline CNN model were developed to precisely identify and classify protected natural areas.

Although these models can effectively detect specific patterns that characterize such areas, the patterns are mostly not inheritably accessible or interpretable by humans. Therefore, explainable machine learning techniques are utilized to explain the designating patterns that drive the model's decision-making process. In the context of analyzing natural areas, \cite{stomberg_exploring_2022} used an inheritable explainable classification network that produces attribution maps that localize patterns characterizing protected natural areas in satellite imagery.
Standard explainable machine learning techniques, such as occlusion sensitivity maps \cite{zeiler_visualizing_2014}, Gradient-based Class Activation Maps\cite{selvaraju_grad-cam_2020}, and DeepLIFT \cite{shrikumar_learning_2019} can identify influential pixels within the image and their significance in deriving the machine learning model's decision. While these methods can provide partial explanations, they lack the capacity to generate complete and valid explanations that accurately characterize natural areas with minimal human interference. For attributions to be considered valid and complete, they must assign high values to the class-discriminative patterns and be consistent with expert opinions \cite{bennetot_greybox_2022,linardatos_explainable_2020}.

Our contribution introduces a novel approach that leverages activation maximization \cite{mahendran_visualizing_2016} and draws inspiration from Cycle Consistent Generative Adversarial Networks (Cycle GANs) \cite{zhu_unpaired_2020}. Our framework, when combined with domain knowledge, offers complete and valid explanations for the patterns that define natural areas with modern human impact. Activation maximization is an explainable machine learning technique that modifies an input image to maximize its classification score produced by CNN. By extending this idea, we utilize activation maximization in our proposed modified Cycle GAN's objective function to generate two images that maximize and minimize the classification score produced by a CNN.
By optimizing the Cycle GAN for maximizing and minimizing the classification score, the two generated images will exhibit slight variations in their color and texture. By comparing these images, an attribution map with complete and valid explanations can be derived that depicts the characteristic patterns of protected natural areas with minimal human interference.
Katzmann et al. \cite{katzmann_explaining_2021} used a comparable approach for explaining medical decisions made by machine learning models. To the best of our knowledge, this work is the first that integrates activation maximization in the Cycle GAN objective function and incorporates domain knowledge to interpret designating patterns contributing to the concept of naturalness in protected natural areas.

\section{Proposed approach}

\begin{figure*}[t]

\centering
\setlength\fboxsep{0pt}
\setlength\fboxrule{0pt}
\fbox{\includegraphics[trim = 0cm 0.6cm 0cm 2.5cm,width=0.65\textwidth]{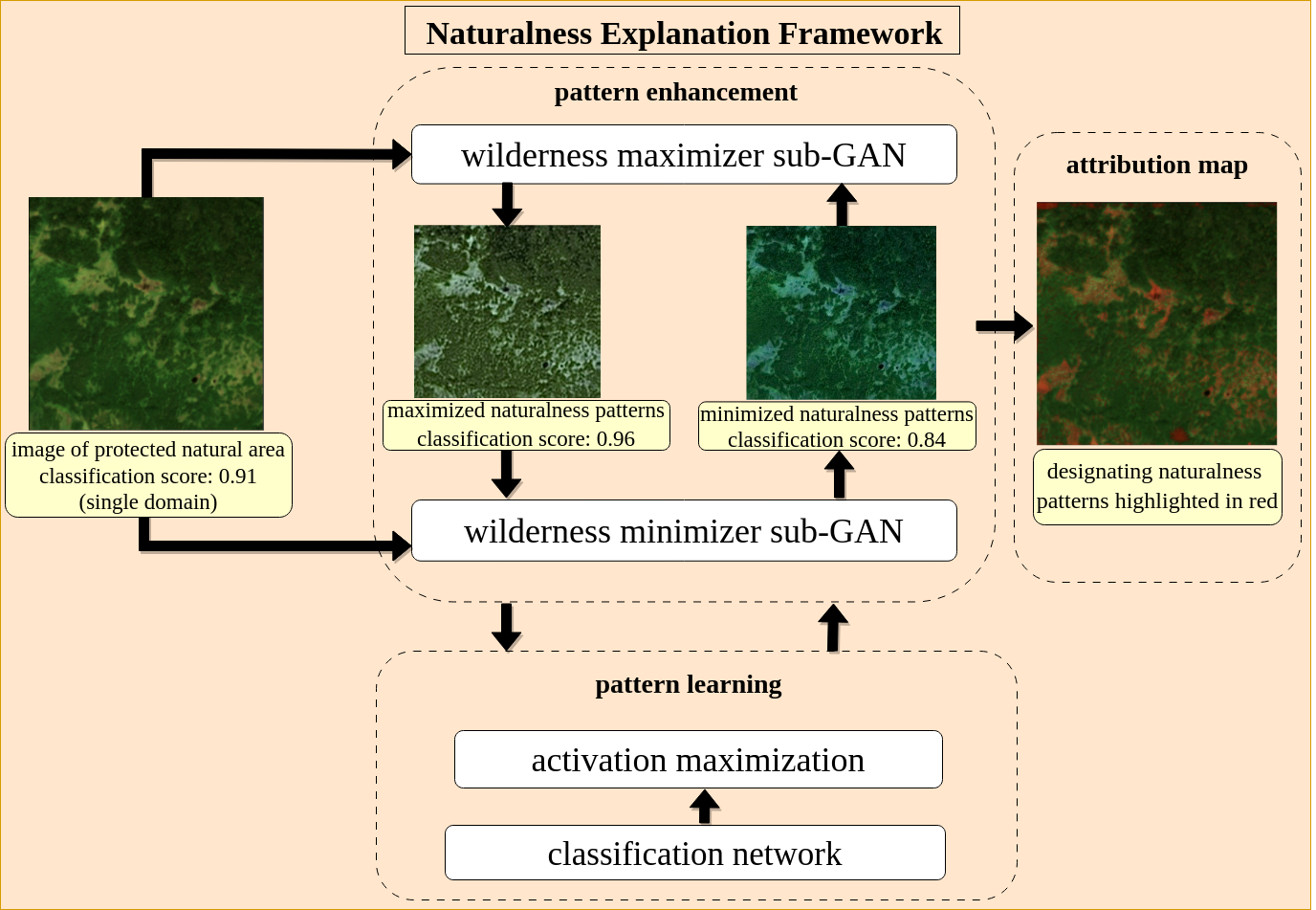}}
\caption{Flowchart of the proposed approach: (left part) The input image displays the class of interest, and it has a classification score of 0.91 from the classification network. (top middle part) pattern enhancement:  Cycle GAN consists of the pattern-maximizer sub-GAN that generates an image with enhanced patterns of the specific class and the pattern-minimizer sub-GAN that generates an image of the same size but with suppressed characteristics. The synthesized images have classification scores of 0.96 and 0.84}, respectively. (bottom middle part) pattern learning: trained classifier for the class of interest being fed with the generated images from b, giving feedback to both generators in the Cycle GAN \cite{zhu_unpaired_2020} to maximize and minimize the characteristics in both sub-GANs, respectively; (right part) attribution mapping:  showing the contribution of each pixel to the class of interest. To generate the attribution map, we subtract each pixel value in the generated image from the input image and calculate the average difference over all channels. We overlayed the original image with the average absolute difference. We've designed a novel framework that modifies the Cycle GAN's objective to generate both maximized and minimized naturalness patterns, enabling us to create more precise attribution maps. To our knowledge, we are pioneering this framework for explaining naturalness patterns in protected areas.
\label{fig:flowchart}
\end{figure*}

Our approach involves three consecutive phases (see Fig. \ref{fig:flowchart}) and is used to analyze and interpret designating patterns of protected natural areas.

\begin{figure}
\centering
\setlength\fboxsep{0pt}
\setlength\fboxrule{0.25pt}
  \includegraphics[trim = 0cm 0.9cm 0cm  0.8cm, width=  0.9\linewidth]{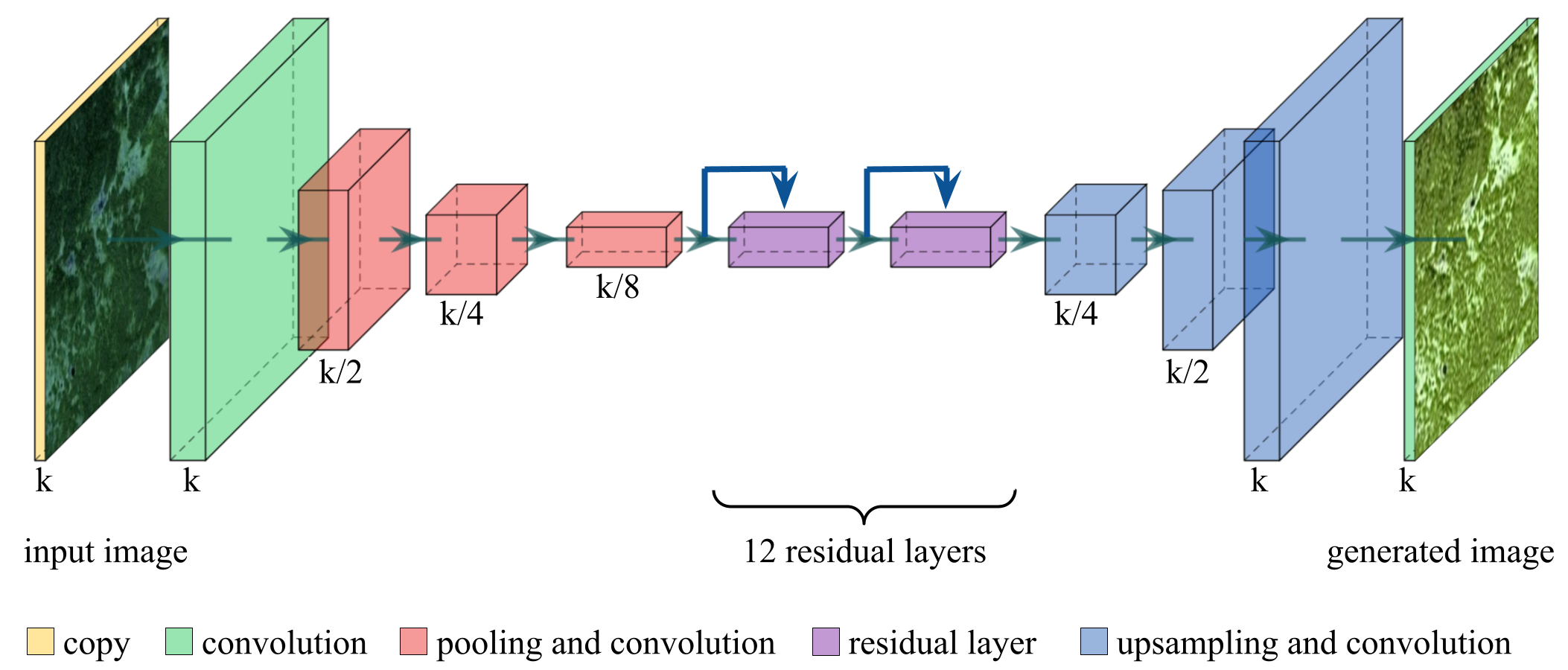}
  \caption{Generator's architecture used in both sub-GANs in the pattern enhancement phase.}
  \label{fig:generator}
\end{figure}

\begin{figure}
  \includegraphics[trim =0cm 0cm 0cm  0cm,width=\linewidth]{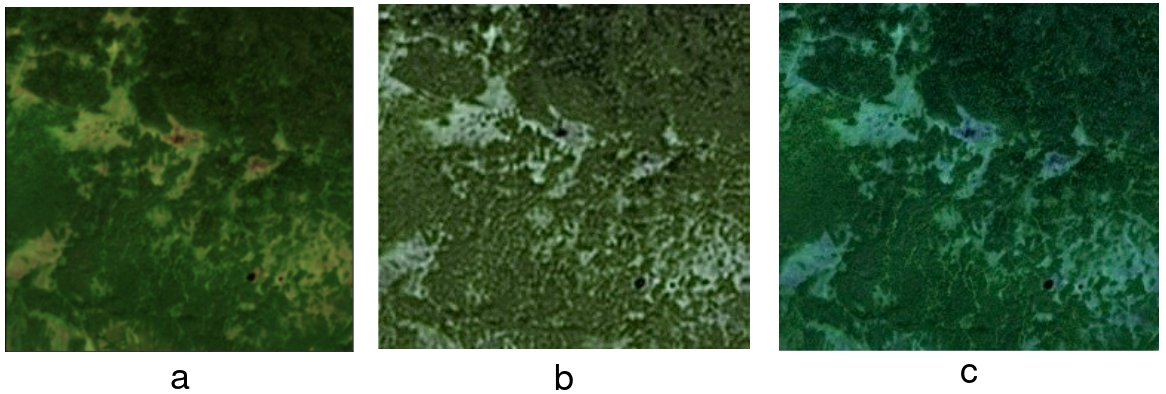}
  \caption{ Generated outputs from the sub-GANs. (a) Original image. (b) the output of the pattern maximizer sub-GAN. (c)the output of the pattern minimizer sub-GAN.
  }
  \label{fig:result1}
\end{figure}

\subsection{Pattern learning}
In the pattern learning phase, we train a single-class classifier to learn the characteristic and distinctive patterns of a class of interest. To enable this, we use a network that has a single output, which can be defined either in the style of a regression network as a score in the range [0, 1], as proposed in \cite{stomberg_exploring_2022} or as a binary decision classifier. 

\subsection{Pattern enhancement}
The pattern enhancement phase is conducted after completing the training process of the classifier in the pattern learning phase.
Here, we integrate the concept of activation maximization in our modified cyclic consistency GAN's objective function to generate two images with pattern-maximized and pattern-minimized image characteristics.
In contrast to commonly used GANs, as the ones proposed by \cite{zhu_unpaired_2020} and \cite{katzmann_explaining_2021} that are primarily designed for performing domain transfer from one image domain to another image domain, our approach has only a single input domain and the generated outputs would be of the same domain as the input's domain with minimum changes pinpointing the important patterns in the input image influencing the classifier's decisions.

\paragraph{Architecture}
We modify the original Cycle GAN's architecture and the objective function, as presented in \cite{zhu_unpaired_2020}, to generate two images based on the input image with minimum changes while keeping the input image's main structures unchanged. These changes lead to the maximization or minimization of the classification score of the generated images.
Our Cycle GAN consists of two sub-GANs, namely the pattern-maximizer and the pattern-minimizer. Each sub-GAN consists of two neural networks, a generator $\ical w$, and a discriminator $\ical d$. Both generators are ResNet-based CNNs\cite{he_deep_2015} with 12 residual blocks. Each generator produces two modified versions of the input image of the same size as the input image. The generator's architecture is illustrated in Fig. \ref{fig:generator}.

\paragraph{Loss function}
The discriminators in both sub-GANs are trained using binary cross entropy:
\begin{equation}\label{eqn:cross entropy}
    \ical{L}_{\text{CE}}(h, \hat{h}) = h\cdot  \log(\hat{h}) + (1 - h) \cdot  \log(1 - \hat{h})\,. 
\end{equation} 
The scalar $h$ is a binary variable indicating whether a sample is real or generated, and $\hat{h}$ is the discriminator’s prediction. 

The overall generator's loss function in both sub-GANs consists of four terms.
The similarity loss term enforces the similarity between the generated images and the original input image:
\begin{equation}\label{eqn:similar}
    \ical{L}_{\text{sim}}(\m{X},\hat{\m{X}}) = |\m{X}- \hat{\m{X}}|_F\,,
\end{equation}
where $\m{X}$ is the input image, and $\hat{\m{X}}$ is the generated pattern-maximized or pattern-minimized output image.
The cycle consistency loss term enforces that successive mapping through both generators leads to a reconstruction close to the original input image:
\begin{equation}\label{eqn:cycle}
    \ical{L}_{\text{cyc}}(\ical{w}^{-},\ical{w}^{+},\m{X}) = |\m{X}- (\ical{w}^{+}(\ical{w}^{-}(\m{X})))|\,,
\end{equation} 
where $\ical{w}^{+}$ is the pattern-maximizing sub-GAN generator, and $\ical{w}^{-}$ is the pattern-minimizing sub-GAN generator.
Moreover, the cycle consistency loss enforces the similarity between the two generated images and the input image. 
The adversarial training loss guides the generator $\ical{w~}$ to generate realistic images following the guidance of the discriminator $\ical{d~}$:
\begin{equation}\label{eqn:gan_gen}
    \ical{L}_{\text{adv}}(\ical{d}_{},\ical{w},\m{X}) = (\ical{d~}(\ical{w~}(\m{X}))-1)^2\,.
\end{equation} 
The activation maximization loss term, as proposed in \cite{nguyen_synthesizing_2016} and \cite{katzmann_explaining_2021}, compels the generator to add features to the generated images that will increase and decrease its classification score for both the pattern-maximizer and pattern-minimizer generators, respectively:
\begin{equation} \label{eqn:AM}
     \ical{L}_{\text{AM}}(y, \ical{w}, \m{X}) =\ical{L}_{\text{CE}}(y, \ical{w~}(\m{X})) \,,
\end{equation}
where $y$ is the label which is 1 for the pattern-maximizer and 0 for the pattern-minimizer, with $\ical{L}_{\text{CE}}$ computed using Eq. \ref{eqn:cross entropy}. 
For both similarity and cycle consistency loss terms, we use the $L_1$ loss function. We use the binary cross entropy loss function for the adversarial and activation maximization loss.

The training of both sub-GANs is performed in two steps. In the pre-training phase, we use the loss term without activation maximization loss term to initialize the weights of both sub-GANs and to ensure that both generators can reconstruct the input image. The main training was performed using the complete generator loss function, which is defined as follows for the pattern-maximizer:
\begin{equation}\label{eqn:generator's loss}
\begin{gathered}[b]
\ical{L}_{\text{max}}(\m{X}, y, \ical{w}^{+}, \ical{w}^{-}, \ical{d}^{+}) = \\  
\ical{L}_{\text{sim}}(\m{X},\hat{\m{X}}) + \ical{L}_{\text{cyc}}(\ical{w}^{-},\ical{w}^{+},\m{X})\\  
\ical{L}_{\text{adv}}(\ical{d}^{+},\ical{w}^{+},\m{X})
+ \lambda \ical{L}_{\text{AM}}(y, \ical{w}^{+}, \m{X}) \,,
\end{gathered}
\end{equation}
where $\lambda$ is the weighting factor assigned to the activation maximization loss term, $\ical{w}^{-}$ being the pattern-minimizing generator, and $ \ical{d}^{-}$ its respective discriminator. 
The loss function is  applied accordingly to the pattern-minimizer generator $\ical{w}^{-}$ by inverting ${(\cdot)}^+$ and ${(\cdot)}^-$ and setting $y=0$. 

\subsection{Attribution mapping}
To produce the attribution maps of the characterizing patterns, we calculate the average absolute difference between the generated images for each image band. Higher attributions are assigned to the areas with a higher average absolute difference because these areas are related to the designating patterns of the class of interest.

\section{Experiments, results, and discussions}
In our use case, we aim to deepen our understanding of the most influential characteristics in the protected natural areas in Fennoscandia (Norway, Sweden, and Finland) by interpreting and explaining the designating patterns that contribute to the naturalness authenticity of this region.

\subsection{Data set and experimental setup}
\subsubsection{AnthroProtect Dataset}
We use the AnthroProtect dataset \cite{stomberg_exploring_2022} consisting of approximately 24,000 multispectral Sentinel-2 images of size $256 \times 256$ pixels, showing protected or anthropogenic areas in Fennoscandia. The protected natural
areas are chosen based on the World Database on Protected
Areas (WDPA) \cite{unep-wcmc_and_iucn_protected_2018}. The protected natural areas are located within the categories "strict nature reserve"~(Ia), "wilderness"~(Ib), and "national park"~(II). The anthropogenic areas are within or close to areas dominated by the CORINE land cover classes "artificial surfaces" and "agricultural areas" while the CORINE dataset represents that domain knowledge \cite{european_environment_agency_corine_2019}. The AnthroProtect dataset thus contains images that lie somewhere at the extremes of the unspecific, continuous wilderness scale within Fennoscandia. To minimize the computational cost, we restrict ourselves to using the red, green, and blue bands; however, our proposed framework is flexible regarding the dimensionality of the input. The AnthroProtect dataset has been published with a proposed data split (training, validation, test), which we adopt accordingly.

In the pattern learning phase, any classification neural network can be employed to learn the designating patterns of naturalness-protected areas. We opted for a modified version of the classification network from \cite{stomberg_exploring_2022}, with only three input channels instead of the ten channels as the network in \cite{stomberg_exploring_2022} to conserve computational resources. The use of fewer input channels led to a minor decrease in the classification accuracy but did not have a negative effect on the performance of the pattern enhancement phase.The training, validation, and test set accuracies for the classification network are 98.7 \%, 97.3 \%, and 99.3 \%, respectively.

During the network's training phase, we adopted the one-cycle learning schedule strategy \cite{smith_super-convergence_2019} with a maximum learning rate set at 0.01. To optimize the model's parameters, we employed gradient descent, incorporating a weight decay of 0.0001. Our training data used a batch size of 32, and to ensure consistency, the pixel values of Sentinel-2 images were normalized to a range between 0 and 1 by dividing them by 10,000.

 We incorporated CutMix \cite{yun_cutmix_2019} during training. This technique allowed us to create synthetic images by mixing portions of areas with human influence and regions devoid of anthropogenic impact. This augmentation aids the classifier in evaluating the naturalness integrity of an image, with a minimum value of 0 representing full anthropogenic influence and a maximum value of 1 denoting completely untouched natural areas.

\subsection{Our Modified Cycle GAN}
We use the PatchGAN \cite{demir_patch-based_2018} for both discriminators. Their goal is to classify overlapping $70\times70$ image patches as either generated or real and to provide feedback to the generator to produce a realistic output. This architecture pushes the discriminator not to focus solely on the artifacts in the generated image but on the general quality of the generated image, which leads to beneficiary feedback to the generator during the Cycle GAN's training process \cite{zhu_unpaired_2020}.
We set $\lambda_{AM}$ = 0.3, resulting in high-resolution outputs with minimum artifacts.
We used the Adam optimizer  \cite{kingma_adam_2017} to train the generative model with the decay of first-order momentum of gradients $\beta_1$ = 0.5, and $\beta_2$ = 0.999.

\subsection{Results and discussion}
\begin{figure*}
\centering
\includegraphics[trim = 0cm 0.7cm 0cm 1cm, width=\linewidth]{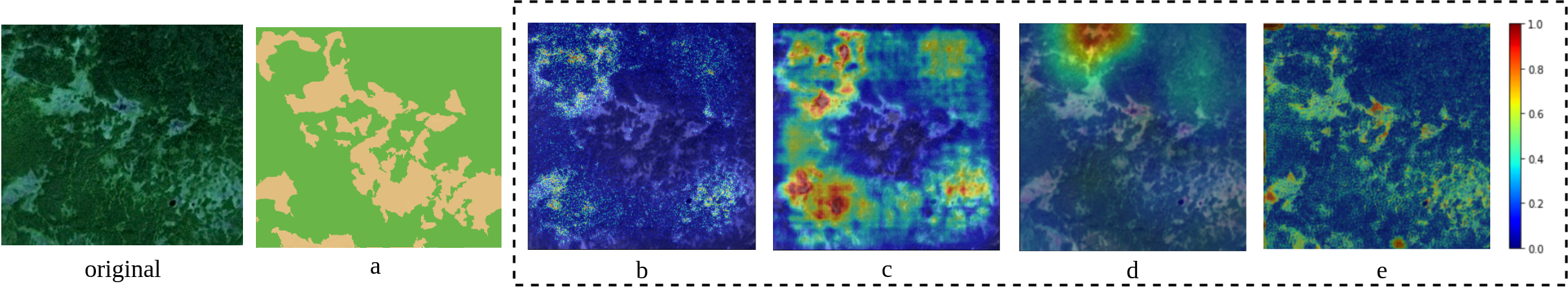}

\caption{Comparison of attribution maps from different explainable machine learning methods. (a) a segmentation mask showing the wetlands pattern in yellow, green representing heavy vegetation, (b) attribution map from DeepLIFT, (c) occlusion sensitivity map, (d) GradCAM, and (e) attribution map from our proposed method. The scale describes plots in the dashed rectangle and assesses the contribution of pixels to the authenticity of protected natural areas. A score of 1 signifies the highest possible contribution. A score of 0 signifies no contribution to the naturalness of protected areas.}

\label{heatmap}
\end{figure*}

\begin{figure}
  \includegraphics[trim = 0cm 0cm 0cm 0 cm, width=\linewidth]{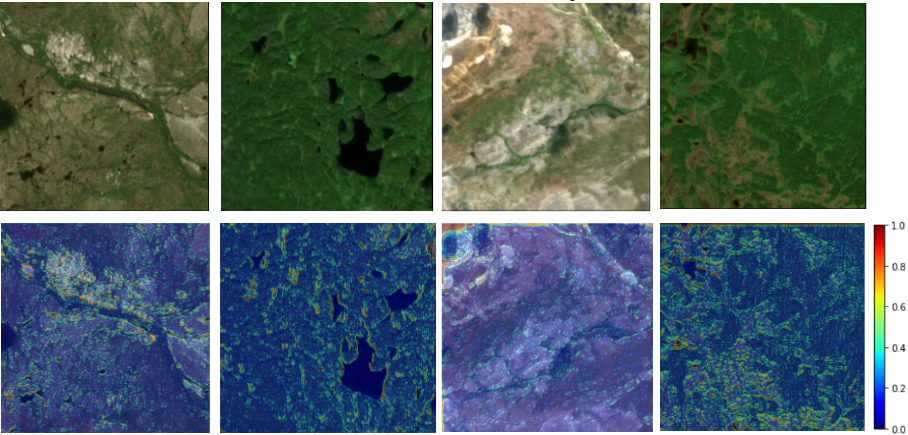}
  \caption{(Top row) shows original input images. (Bottom row) shows the corresponding attribution maps produced by our approach.}
  \label{fig:emam5}
\end{figure}
Figures \ref{fig:flowchart} and \ref{fig:result1} show example images generated by the pattern-maximizer and the pattern-minimizer sub-GANs. 
The pattern-maximizer produces images with higher pixel values in the green channel compared to the original image, whereas the pattern-minimizer produces the same effect on the blue channel. Including environmental domain knowledge and information from existing land cover products, we can see that brighter wetlands areas were produced in the pattern-maximized images, and the borders of wetlands and the surroundings of water bodies are highlighted with vibrant green color by the pattern-minimizer. 

We also depict attribution maps showing the importance of each pixel in the input image, contributing to the naturalness integrity and its influence on the classifier's decision, see Fig. \ref{heatmap}.  
By incorporating our method's attribution map with the correspondent segmentation mask from the CORINE dataset, we can conclude that our method assigns high importance to wetlands and the small water bodies (bonds) in the middle of the image and lower attributions to the non-discriminative vegetation areas between the characterizing features. The occlusion sensitivity map \cite{uchiyama_visually_2022} assigns high attribution to the same features with a lower accuracy but assigns high attributions to non-discriminative vegetation areas between the discriminative patterns. The attribute map produced by the DeepLIFT \cite{shrikumar_learning_2019} does not assign comparable attribution for the wetlands in the middle of the image but can detect them near its borders. GradCAM \cite{selvaraju_grad-cam_2020} applied to the last convolution layer assigns high attribution only to the image's upper middle part with low vegetation. 
Integrating our approach with domain knowledge enables our approach to explain the designating patterns contributing to the concept of wilderness. Attribution maps shown in Figures~\ref{heatmap} and~\ref{fig:emam5} prioritize bare lands, wetlands, and mountain peaks' glaciers, aligning with naturalness characteristics in protected areas in Fennoscandia \cite{gunnarsson_swedish_2014}. This integration provides a more comprehensive explanation of the classification network's decision-making in the pattern learning phase. Additionally, IoU results shown in table \ref{tab:iou} for high attribution pixels, including naturalness patterns like bare lands and wetlands, further validate the effectiveness of our integrated approach.
\begin{table}[ht]
\centering
\begin{tabular}{l|rrrrr}
    $\;\;$ Method $\rightarrow$   & DeepLIFT & OSM & GraddCAM& Ours \\ \hline
     IoU \% &     81.2           &  69.1             & 53.3  & 93.2        
    
\end{tabular}
\caption{{Average intersection over Union (IoU) for the pixels with high attributions with wetlands and bare lands classes.}
\label{tab:iou}}
\end{table}



A combination of wetlands and bare lands have a unique combination of water, soil, and vegetation, given the fact that they are characterized by the presence of waterlogged soils, which are saturated or flooded for most of the year, which hinders the accessibility to the area and minimize anthropogenic impact. These patterns are recognized in domain knowledge as strongly associated with the naturalness of protected areas in Fennoscandia \cite{gunnarsson_swedish_2014}. We can conclude that our method, combined with domain knowledge, produces complete and valid explanations that align with the expert's opinions.

Although our approach introduces two challenges, namely the high computational cost and the possibility of generating artifacts in the generated images \cite{aggarwal_generative_2021}, our approach provides valid and complete explanations by consistently assigning high attributions to characterize geo-ecological patterns such as wetlands.

\section{Conclusion}
We proposed a novel, explainable machine learning approach to suppress and enhance image patterns and produce attribution maps incorporated with domain knowledge to produce valid and complete explanations. 
We utilize the concept of activation maximization and integrate it into a Cycle Consistent Generative Adversarial  Network's objective function to maximize the characterizing patterns of a specific class. 
We utilized our approach to analyze patterns in satellite imagery to better understand natural protected areas. 
We could show that our approach is more suitable to explain the patterns corresponding to the naturalness integrity of the protected natural areas in Fennoscandia. Our proposed method can produce complete, valid explanations of the designating geo-ecological patterns that uphold the ecological authenticity of the protected areas more than other methods.
Since our approach can be extended to multi-class classifications and various applications, we consider it a promising direction in the Earth sciences.

\section*{Acknowledgment}

We acknowledge funding from DFG projects RO~4839/5-1, SCHM~3322/4-1, and RO~4839/6-1, as well as support from DFG's Excellence Strategy, EXC-2070 - 390732324 - PhenoRob.

\ifCLASSOPTIONcaptionsoff
  \newpage
\fi

\bibliographystyle{ieeetr}
\bibliography{ref}
\end{document}